\documentclass{svjour3}
\pdfoutput=1  
\usepackage{anyfontsize}
\usepackage{setspace}
\usepackage{silence}
\usepackage{cite}
\usepackage{amsmath}
\usepackage{amssymb}
\usepackage{enumerate}
\usepackage{subcaption}
\usepackage[table,xcdraw]{xcolor}
\usepackage{graphicx}
\usepackage{algorithm,algorithmic}
\usepackage{bigstrut}
\usepackage{multirow}
\usepackage{bm}
\usepackage{epstopdf}
\usepackage{csvsimple}
\usepackage{float}
\usepackage{threeparttable}
\usepackage{adjustbox}
\usepackage{url}
\usepackage[many]{tcolorbox}
\DeclareUrlCommand\ULurl{}
\usepackage{rotating}
\mathchardef\mhyphen="2D 
\let\oldtheta\theta
\renewcommand{\theta}{\ensuremath{\oldtheta}}
\usepackage{pifont}

\begin{document}
	
	\title{Graph-Embedded Multi-layer \textcolor{red}{Kernel Extreme Learning Machine} for One-class Classification \\or\\Graph-Embedded Multi-layer \textcolor{red}{Kernel Ridge Regression} for One-class Classification \\or\\Graph-Embedded Multi-layer \textcolor{red}{Least Square SVM with zero bias} for One-class Classification}

\author{Chandan Gautam \and 
	Aruna Tiwari \and 
	M. Tanveer
}

\institute{C. Gautam \at
	Indian Institute of Technology Indore, India \\
	\email{chandangautam31@gmail.com}           
	\and
	A. Tiwari \at
	Indian Institute of Technology Indore, India \\
	\email{artiwari@iiti.ac.in}           
	\and
	M. Tanveer \at
	Indian Institute of Technology Indore, India \\
	\email{mtanveer@iiti.ac.in} 		
}

\maketitle

\begin{tcolorbox}[width=7in,colback=white]
	\textcolor{red}{Why three titles?} \textcolor{blue}{Because three methods viz; Kernel ridge regression (KRR), lease square support vector machine with zero bias (LSSVM(bias=0)) and kernel extreme learning machine (KELM), are identical in outcomes and developed by three different researchers under three different framework. Since, KRR are more genric name compared to others, we use name KRR instead of LSSVM or KELM in this paper. Proposed methods of this paper can be considered as variants of KRR or LSSVM(with bias=0) or KELM:}
	
	\centering {\textcolor{red}{\textbf{KELM = KRR = LSSVM(with bias=0)}}}
\end{tcolorbox}

\begin{abstract}
	
\textbf{Introduction:}
A brain can detect outlier just by using only normal samples. Similarly, one-class classification ($OCC$) also uses only normal samples to train the model and trained model can be used for outlier detection.

\noindent
\textbf{Proposed Method:} In this paper, a multi-layer architecture for $OCC$ is proposed by stacking various Graph-Embedded Kernel Ridge Regression ($KRR$) based Auto-Encoders in a hierarchical fashion. These Auto-Encoders are formulated under two types of Graph-Embedding, namely, local and global variance-based embedding. This Graph-Embedding explores the relationship between samples and multi-layers of Auto-Encoder project the input features into new feature space. The last layer of this proposed architecture is Graph-Embedded regression-based one-class classifier. The Auto-Encoders use an unsupervised approach of learning and the final layer uses semi-supervised (trained by only positive samples and obtained closed-form solution) approach to learning. 

\noindent
\textbf{Experimental Results:} The proposed method is experimentally evaluated on $21$ publicly available benchmark datasets. Experimental results verify the effectiveness of the proposed one-class classifiers over $11$ existing state-of-the-art kernel-based one-class classifiers. Friedman test is also performed to verify the statistical significance of the claim of the superiority of the proposed one-class classifiers over the existing state-of-the-art methods.

\noindent
\textbf{Conclusion:} By using two types of Graph-Embedding, $4$ variants of Graph-Embedded multi-layer $KRR$-based one-class classifier has been presented in this paper. All $4$ variants performed better than the existing one-class classifiers in terms of various discussed criteria in this paper. Hence, it can be viable alternative for $OCC$ task. In future, various other types of Auto-Encoders can be explored within proposed architecture.   
\keywords{One-Class Classification \and Outlier Detection \and Kernel Ridge Regression \and Graph-Embedding \and Multi-layer}	
\end{abstract}

\section{Introduction}
\textbf{O}ne-\textbf{c}lass \textbf{C}lassification (OCC) has been widely used for outlier, novelty, fault, and intrusion detection \cite{moya1993one,khan2009survey,pimentel2014review,xu2013rough,hamidzadeh2018improved,xiao2009multi} by researchers from different disciplines. In multi-class problem, both positive and negative samples are available for training \cite{gepperth2016generative,luria2014detection,justodetection,anbar2018machine}. However, in OCC problems, samples of the class of interest (i.e., positive samples) are available while negative samples are very rare or costly to collect \cite{david2001tax,park2007svdd,liu2013svdd,kassab2009incremental,munoz2006estimation,chen2017one,hu2015privacy}, thus making the application of multi-class models problematic. Various one-class classifiers \cite{pimentel2014review,o2014anomaly} have been proposed based on the regression model, the clustering model etc. One-class classification methods available in the literature can be divided into two broad categories viz., non-kernel-based and kernel-based methods. Various non-kernel-based one-class classifiers are principal component analysis based data descriptor\footnote{One-class classifiers are also known as data descriptors due to their capability to describe the distribution of data and the boundaries of the class of interest} \cite{david2001tax}, angle-based outlier factor data description \cite{kriegel2008angle}, K-means data description \cite{david2001tax}, self-organizing map data description \cite{david2001tax}, Auto-Encoder data descriptor \cite{japkowicz1999concept} etc. Whereas, the kernel-based one-class classifiers are support vector data description \cite{tax2004support}, one-class support vector machine\cite{scholkopf1999support}, kernel principal component analysis based data description \cite{hoffmann2007kernel} etc. However, kernel-based methods have been shown to outperform non-kernel-based methods in the literature \cite{pimentel2014review,david2001tax}. Despite this fact, these kernel-based methods involve the solution of a quadratic optimization problem, which is computationally expensive. Apart of these kernel-based methods, $KRR$-based models \cite{saunders1998ridge} optimize the problem rapidly in a non-iterative way by solving a linear systems. Therefore, $KRR$-based models \cite{saunders1998ridge,wornyo2018co,zhang2017benchmarking,he2014kernel,wu2017cost} have received quite attention by researchers for solving various types of problems viz., regression, binary, multi-class etc.

In recent years, various $KRR$-based\footnote{\label{kelmkrr}Methods discussed in this paragraph have used name KELM in their paper. Since, KELM and KRR are identical as discussed in the above paragraph, we use more generic name KRR instead of KELM.} one-class classifiers have been developed and exhibited better performance compared to various state-of-the-art one-class classifiers. Overall, the $KRR$-based one-class classifiers can be divided into two types, namely, (i) without Graph-Embedding (ii) with Graph-Embedding. For `without Graph-Embedding', two types of architectures have been explored for OCC. One is $KRR$-based single output node architecture \cite{leng2014one}\textsuperscript{\ref{kelmkrr}}, and other is $KRR$-based Auto-Encoder architecture \cite{gautam2017construction}\textsuperscript{\ref{kelmkrr}}. For `with Graph-Embedding', Iosifidis et al.\cite{iosifidis2016one}\textsuperscript{\ref{kelmkrr}} presented local and global variance-based Graph-Embedded one-class classifier. Different types of Laplacian Graphs are employed by Iosifidis et al.\cite{iosifidis2016one} for local (i.e., Local Linear Embedding, Laplacian Eigenmaps etc.) and global (linear discriminant analysis and clustering-based discriminant analysis etc.) variance embedding. Later, global variance-based Graph-Embedding has been extended in order to exploit class variance and sub-class variance information for face verification task by Mygdalis et al.\cite{mygdalis2016one}\textsuperscript{\ref{kelmkrr}}. All the above-mentioned $KRR$-based one-class classifiers employ only single-layered architecture.

Over the last decade, stacked Auto-encoder based multi-layer architectures have received quite attention by researchers for multi-class and binary class classification tasks \cite{bengio2009learning,schmidhuber2015deep}. Such architectures can lead to better representation learning \cite{vincent2008extracting,shin2013stacked} and also used in dimensionality reduction \cite{hinton2006reducing,van2009dimensionality,wang2014generalized}. High-level feature representations obtained by using stacked Auto-Encoder also helps in improving the performance of the traditional classifiers \cite{vincent2010stacked}. This paper explores the possibility of $KRR$-based representation learning using stacked Auto-Encoder for the one-class classification task.

In this paper, we propose a multi-layer architecture by stacking various Graph Embedded $KRR$-based Auto-Encoders (trained using unsupervised learning) in a hierarchical manner for one-class classification task. These Auto-Encoders are designed to exploit two types of data relationships encoded in graphs \cite{yan2007graph}, i.e. local and global variance information-based Graph-Embedding. These information are incorporated in the Auto-Encoder training process in order to simultaneously enhance the data reconstruction ability, data representation ability, and the class compactness in the derived feature space. The multiple layers exploit the idea of successive nonlinear data mappings and hence capture the relationship effectively. After stacking several Auto-Encoder layers in a hierarchical manner, data are represented in a new feature space in which Graph-Embedded regression-based one-class classifier is employed in the final layer. At final layer, output of the stacked Auto-Encoder is approximated to any real number and set a threshold for deciding whether any sample is outlier or not. Two types of threshold deciding criteria (i.e. $\theta1$ and $\theta2$) are discussed so far in this paper. By employing different realizations of the proposed Auto-Encoder, two different architectures are formed based on the local and global variance criteria and are referred as $LMKOC$ and $GMKOC$, respectively. Both architectures are experimented with two types of threshold criteria and developed 4 variants of the Graph-Embedded multi-layer one-class classifier. Further, the performance of $GMKOC$ and $LMKOC$ are evaluated using $21$ benchmark datasets and its performance is compared with $11$ state-of-the-art kernel-based methods available in the literature. Finally, a Friedman test \cite{demvsar2006statistical} is conducted to verify the statistical significance of the experimental outcomes of the proposed classifiers and it rejects the null hypothesis with $95\%$ confidence level. 


The rest of the paper is organized as follows. Section \ref{Sec:proposed} describes the $LMKOC$ and $GMKOC$ in detail. Performance evaluation is provided in Section \ref{Sec:performance}. Finally, Section \ref{Concl} concludes our work.

\section{Proposed Method}\label{Sec:proposed}
\begin{figure*}[t]
	\begin{center}
		\scalebox{0.48}{\includegraphics{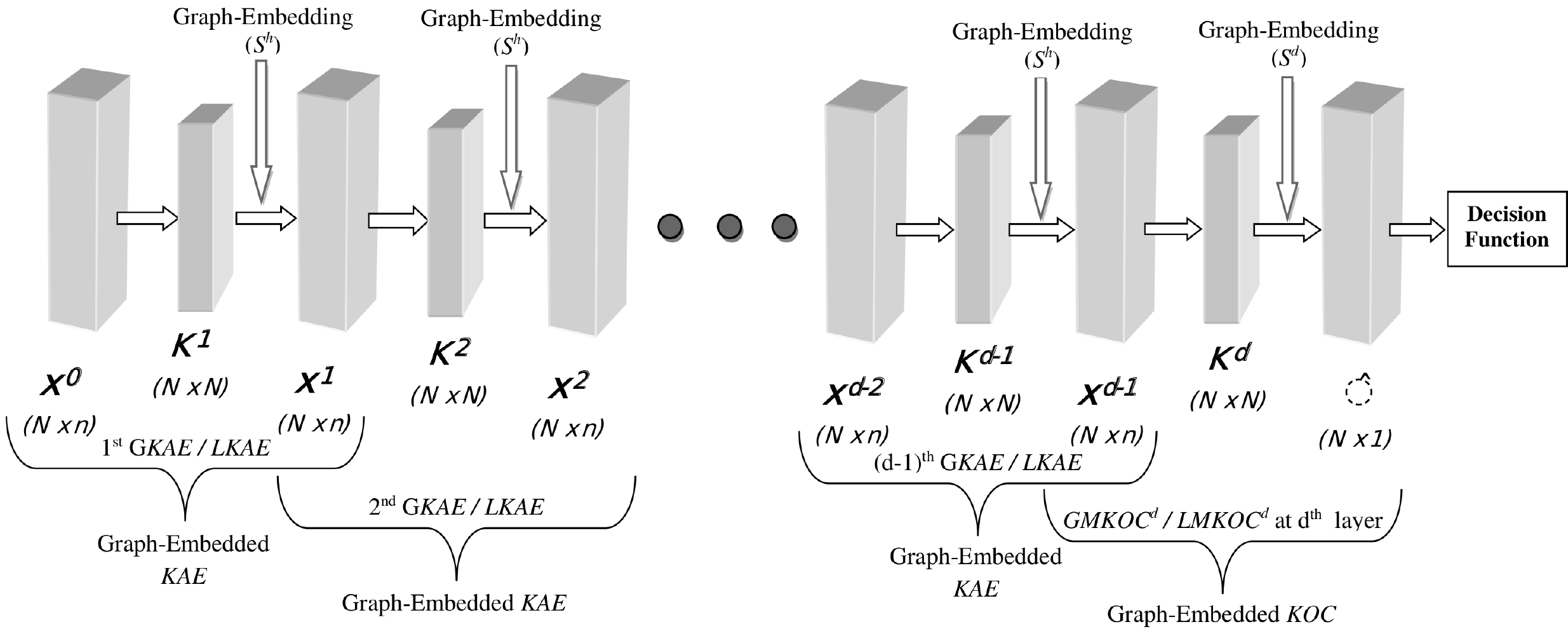}}
		\caption{Schematic Diagram of Graph-Embedded multi-layer KRR-based architecture for one-class classification}
		\label{fig:mlocelm}
	\end{center}
\end{figure*}

In this section, a Graph-Embedded multi-layer KRR-based architecture for one-class classification is described. The proposed multi-layer architecture is constructed by stacking various Graph-Embedded $KRR$-based Auto-Encoders, followed by a Graph-Embedded $KRR$-based one-class classifier, as shown in Fig. \ref{fig:mlocelm}. Graph-Embedding is performed by two types of variances information viz., local and global variance. One is referred as \textbf{L}ocal variance based Graph-Embedded \textbf{M}ulti-layer $\bm{K}RR$ for \textbf{O}ne-class \textbf{C}lassification ($\bm{LMKOC}$), and other is referred as \textbf{G}lobal variance based Graph-Embedded \textbf{M}ulti-layer $\bm{K}RR$ for \textbf{O}ne-class \textbf{C}lassification ($\bm{GMKOC}$). \textbf{L}ocal and \textbf{g}lobal variance-based \textbf{k}ernelized \textbf{A}uto-\textbf{E}ncoders are referred as $LKAE$ and $GKAE$, respectively.  

During construction of multi-layer architecture, use either local or global variance for every layers of the architecture. As shown in  Fig. \ref{fig:mlocelm}, $GMKOC$/$LMKOC$ is constructed by stacking various $GKAE$s/$LKAE$s\footnote{Here, '/' denotes or. $GMKOC$ uses $GKAE$ and $LMKOC$ uses $LKAE$.}. These stacked Auto-Encoders are employed for defining the successive data representation. In the $1^{st}$ $GKAE$/$LKAE$ of this figure, input training matrix is denoted by $\bm{X=X^0=\left\{x_i^0\right\}}$, where $\bm{x_i^0}=[x_{i1}^{0}, x_{i2}^{0},...,x_{in}^{0}]$, $i=1,2,...,N$, is the $n$-dimensional input vector of the $i^{th}$ training sample. Let us assume that there are $d$ layers in the proposed architecture, i.e., $h=1, 2,...,d$. Output of the $h^{th}$ layer is passed as input to the $(h+1)^{th}$ layer. Let us denote output at $h^{th}$ layer of Auto-Encoder, $\bm{X^h=\left\{x_i^h\right\}}$, where $\bm{x_i^h}=[x_{i1}^{h}, x_{i2}^{h},...,x_{in}^{h}]$, $i=1,2,...,N$. $\bm{X^h}$ corresponds to the output of the $h^{th}$ Auto-Encoder and the input of the $(h+1)^{th}$ Auto-Encoder. Each of the Auto-Encoders involves a data mapping using function $\phi(.)$, mapping $\bm{X^{h-1}}$ to $\bm{\phi^h=\phi(X^{h-1})}$. $\phi(.)$ corresponds to a mapping of $\bm{X^{h-1}}$ to the corresponding kernel space $\bm{K^h}=\bm{(\Phi^h)^T\Phi^h}$. Here, $\bm{\Phi^h}=\left [\bm{\phi_1^h, \phi_2^h,...,\phi_N^h\right ]}$. The data representation obtained by calculating the output of the $(d-1)^{th}$ Auto-Encoder in the architecture is passed to the $d^{th}$ layer for OCC using $GMKOC^d$/$LMKOC^d$. Here, $GMKOC^d$/$LMKOC^d$ denotes $d^{th}$ layer of $GMKOC$/$LMKOC$. In the give figure, Graph-Embedding is performed by using a scattered matrix $\bm{S^h}$, which encodes the local or global variance information with the kernel matrix. Here, $\bm{S^h}$ denotes scattered matrix of $h^{th}$ layer. Two types of training errors and weight matrices are generated by $GMKOC$/$LMKOC$. The first type of training error matrix and weight matrix are generated by the $h^{th}$ Auto-Encoder until $(d-1)$ layers and denoted as $\bm{ E^h=\left\{e_i^h\right\}}$ and $\bm{\beta_a^h}$, where $i=1, 2,...,N$ and $h=1, 2,...,(d-1)$, respectively. And the other type of training error vector and weight vector are generated by the one-class classifier at $d^{th}$ layer and denoted as $\bm{E^d}=\left\{e_i^d\right\}$ and $\bm{\beta_o^d}$, where $i=1, 2,...,N$, respectively.
Based on the above notations, proposed methods $GMKOC$ and $LMKOC$ are discussed in the next subsections.

\subsection{Local Variance Information based Graph-Embedded Multi-layer $KRR$ for One-class Classification: $LMKOC$}\label{subsec:ML-GOCKELM}
In this subsection, $LMKOC$ is proposed. This multi-layer architecture exploits \textbf{L}ocal variance information with $\bm{KAE}$s ($LKAE$s). The overall architecture of $LMKOC$ is formed by two processing steps.

In \textbf{first step}, $(d-1)$ $LKAE$s are trained, each defining a triplet ($\bm{X^h, \beta_a^h, S^h}$), and stacked in a hierarchical manner. A $LKAE$ involves non-linear mapping $\bm{X^{h-1} \rightarrow \Phi^h}$ and, subsequently, defines a graph $\mathcal{G}^h=\bm{\left\{\Phi^h,V^h\right\}}$ where $\bm{V^h} \in \mathbb{R}^{N \times N}$ is the weight matrix expressing similarities between the graph nodes $\bm{\phi_}i\bm{^h}\in\Phi^h$. The Graph Laplacian matrix of the $h^{th}$ $LKAE$ is calculated by $\bm{\mathcal{L}^h=D^h-V^h}$, where $\bm{D^h}$ is a diagonal degree matrix in the $h^{th}$ layer defined as \cite{yan2007graph}:
 \begin{equation}
   \label{Eq:Diagonal}
   \begin{aligned}
   \bm{D_{ii}^h}=\sum_{j=1}^{N}\bm{V_{ij}^h}
   \end{aligned}
 \end{equation}
Any type of local variance based Laplacian Graph (e.g. Laplacian Eigenmaps ($LE$) \cite{belkin2003laplacian}, Locally Linear Embedding ($LLE$) \cite{saul2003think} etc.) can be exploited in the $LKAE$. In our experiments, we have used the fully connected and k-nearest neighbor graph using the heat kernel function:
\begin{equation}
\label{Eq:pairwise_similarity}
\begin{aligned}
\bm{v_{ij}^h}= exp\left (-\frac{\left \|\bm{ \phi_i^h}-\bm{\phi_j^h} \right \|^2_2}{2\sigma^2} \right )
\end{aligned}
\end{equation}
where, $\sigma$ is a hyper-parameter scaling the square Euclidean distance between $\bm{\phi_i^h}$ and $\bm{\phi_j^h}$. In the case of k-nearest neighbor Graph, the weight matrix $\bm{V^h}$ is defined as follows:
\begin{equation}
\label{Eq:knn_V_lin}
\begin{aligned}
\bm{V_{ij}^h}=\left\{\begin{matrix}
v_{ij}^h, & \text{if} \ \bm{\phi_j^h}\in\mathcal{N}_i^h\\
0, & \text{otherwise}
\end{matrix}\right.
\end{aligned}
\end{equation}
where, $\mathcal{N}_i^h$ denotes the neighborhood of $\bm{\phi_i^h}$. Using the above notation, the scatter matrix $\bm{S^h}$ encoding the local variance information is given by:
\begin{equation}
\label{Eq:S_scatter}
\begin{aligned}
\bm{S^h} = \bm{\Phi^h\mathcal{L}^h(\Phi^h)^T}
\end{aligned}
\end{equation}

Minimization criterion of $LKAE$ is derived by using vanilla $KRR$-based Auto-Encoder ($KAE$). A $KAE$ can be formulated as follows:

\begin{equation}
\label{Eq:trans_mlaaelm}
\begin{aligned}
\text{Minimize}:\pounds_{KAE}=\frac{1}{2}\left \|\bm{\beta_a^h}\right\|^{2} + \frac{C}{2}\sum_{i=1}^{N}\left\|\bm{e_i^h}  \right \|_2^{2}  \\
\text{Subject to}:\ \bm{(\beta_a^h)^T \phi_i^h =x_i^{h-1} - e_i^h}, \text{  }i=1,2,...,N,
\end{aligned}
\end{equation}
where $C$ is a regularization parameter, and $\bm{e_i^h}$ is a training error vector corresponding to the $i^{th}$ training sample at $h^{th}$ layer. Based on the minimization criterion in (\ref{Eq:trans_mlaaelm}), $LKAE$ can be formulated as follows:

\begin{equation}
\label{Eq:GELM-AE}
\begin{aligned}
\text{Minimize}:\pounds_{LKAE}=\frac{1}{2} Tr\Big(\bm{(\beta_a^h)^T(S^h}+\lambda \bm{I)\beta_a^h} \Big) + \frac{C}{2}\sum_{i=1}^{N}\left\|\bm{e_i^h}  \right \|_2^{2}  \\
\text{Subject to}:\ \bm{(\beta_a^h)^T \phi_i^h=x_i^{h-1} - e_i^h}, \text{  }i=1,2,...,N,
\end{aligned}
\end{equation}

Based on the Representer Theorem \cite{argyriou2009there}, we express $\bm{\beta^h_a}$ as a linear combination of the training data representation $\bm{\Phi^h}$ and a reconstruction weight matrix $\bm{W_a^h}$:
\begin{equation}
\label{Eq:representation_theorem}
\begin{aligned}
\bm{\beta^h_a} = \bm{\Phi^h W_a^h}.
\end{aligned}
\end{equation}
Hence, by using Representer Theorem \cite{argyriou2009there}, minimization criterion in (\ref{Eq:GELM-AE}) is reformulated as follows:
	\begin{equation}\label{Eq:GKELM-AE1}
	\begin{aligned}
	\text{Minimize}:\pounds_{LKAE}=\frac{1}{2}  Tr\Big(\bm{(W_a^h)^T(\Phi^h)^T(\Phi^h\mathcal{L}^h(\Phi^h)^T} \\ +\lambda \bm{I)\Phi^hW_a^h}\Big) + \frac{C}{2}\sum_{i=1}^{N}\left\|\bm{e_i^h}  \right \|_2^{2},  \\
	\text{Subject to}:\ \bm{(W_a^h)^T(\phi_i^h)^T\phi_i^h=x_i^{h-1} - e_i^h}, \text{  }i=1,2,...,N.
	\end{aligned}
	\end{equation}
By further substitution of $\bm{K^h}=\bm{(\Phi^h)^T\Phi^h}$, where $\bm{k_i^h}\subseteq \bm{K^h}$ is formed by the elements $\bm{k_{ij}^h}=\bm{(\phi_i^h)^T\phi_j^h}$, the criterion in (\ref{Eq:GKELM-AE1}) can be written as:	
	\begin{equation}\label{Eq:GKELM-AE2}
	\begin{aligned}
	\text{Minimize}:\pounds_{LKAE}=\frac{1}{2} Tr\Big(\bm{(W_a^h)^T(K^h\mathcal{L}^h K^h}+\lambda \bm{K^h)W_a^h}\Big)\\ + \frac{C}{2}\sum_{i=1}^{N}\left\|\bm{e_i^h}  \right \|_2^{2},  \\
	\text{Subject to}:\ \bm{(W_a^h)^T k_i^h = x_i^{h-1} - e_i^h}, \text{  }i=1,2,...,N.
	\end{aligned}
	\end{equation}
 The Lagrangian relaxation of (\ref{Eq:GKELM-AE2}) is shown below in (\ref{Eq:GKELM-AE2_lang}):	
	\begin{equation}\label{Eq:GKELM-AE2_lang}
	\begin{aligned}
	\pounds_{LKAE}=\frac{1}{2} Tr\Big(\bm{(W_a^h)^T(K^h\mathcal{L}^h K^h}+\lambda \bm{K^h)W_a^h}\Big) \\ + \frac{C}{2}\sum_{i=1}^{N}\left\|\bm{e_i^h}  \right \|_2^{2} - \sum_{i=1}^{N}\bm{\alpha_i^h}(\bm{(W_a^h)^T k_i^h - x_i^{h-1} + e_i^h})
	\end{aligned}
	\end{equation}
where $\bm{\alpha^h = \{\alpha_i^h\}}, i=1,2 \hdots N$, is a Lagrangian multiplier. In order to optimize (\ref{Eq:GKELM-AE2_lang}), we compute its derivatives as follows:
\begin{equation}
 \label{Eq:GKELM-AE2_deriv1}
 \begin{aligned}
 &\frac{\partial \pounds_{LKAE}}{\partial \bm{W_a^h}} = 0 \Rightarrow \bm{W_a^h}=(\bm{\mathcal{L}^h K^h}+\lambda\bm{I})^{-1}\bm{\alpha^h}
  \end{aligned}
\end{equation}
\begin{equation}
 \label{Eq:GKELM-AE2_deriv2}
 \begin{aligned}
 &\frac{\partial \pounds_{LKAE}}{\partial \bm{e_i^h}} = 0 \Rightarrow \bm{E^h}=\frac{1}{C}\bm{\alpha^h}
  \end{aligned}
\end{equation}
\begin{equation}
 \label{Eq:GKELM-AE2_deriv3}
 \begin{aligned}
 &\frac{\partial \pounds_{LKAE}}{\partial \bm{\alpha_i^h}} = 0 \Rightarrow \bm{(W_a^h)^T K^h = X^{h-1} - E^h} \\
 \end{aligned}
\end{equation}
The matrix $\bm{W^h_a}$ is obtained by substituting (\ref{Eq:GKELM-AE2_deriv2}) and (\ref{Eq:GKELM-AE2_deriv3}) into (\ref{Eq:GKELM-AE2_deriv1}), and is given by:
\begin{equation}
\label{Eq:W_GKELM-AE}
\begin{aligned}
\bm{W^h_a} &= \left(\bm{K^h}+\frac{1}{C}\bm{\mathcal{L}^hK^h}+\frac{\lambda}{C}\bm{I}\right)^{-1}\bm{X^{h-1}},
\end{aligned}
\end{equation}
Now, $\bm{\beta^h_a}$ can be derived by substituting (\ref{Eq:W_GKELM-AE}) into (\ref{Eq:representation_theorem}):
\begin{equation}
\label{Eq:ow_GKELM-AE}
\begin{aligned}
\bm{\beta^h_a} &= \bm{\Phi^h}\left(\bm{K^h}+\frac{1}{C}\bm{\mathcal{L}^hK^h}+\frac{\lambda}{C}\bm{I}\right)^{-1}\bm{X^{h-1}}.
\end{aligned}
\end{equation}

After mapping the training data through the $(d-1)$ successive $LKAE$s in the \textbf{first step}, the training data representations defined by the outputs of the $(d-1)^{th}$ $LKAE$ are used in order to train a \textbf{L}ocal variance based Graph-Embedded \textbf{M}ulti-layer \textbf{K}RR for \textbf{O}C\textbf{C} at $d^{th}$ layer ($LMKOC^d$) in the \textbf{second step}. The $LMKOC^d$ involves a nonlinear mapping $\bm{X^{d-1} \rightarrow \Phi^d}$ and is trained by solving the following optimization problem:
\begin{equation}
\label{Eq:ML-GOCELM}
\begin{aligned}
\text{Minimize}:\pounds_{LMKOC^d}=\frac{1}{2}\bm{(\beta_o^d)^T(S^d+\lambda I)\beta_o^d} + \frac{C}{2}\sum_{i=1}^{N}\left\|e_i^d  \right \|_2^{2}  \\
\text{Subject to}:\ \bm{(\beta_o^d)^T \phi_i^d}=r - e_i^d, \text{  }i=1,2,...,N,
\end{aligned}
\end{equation}

By using Representer Theorem \cite{argyriou2009there}, $\bm{\beta^d_o}$ is expressed as a linear combination of the training data representation $\bm{\Phi^d}$ and reconstruction \textbf{weight vector} $\bm{W_o^d}$:
\begin{equation}
\label{Eq:representation_theorem_oc}
\begin{aligned}
\bm{\beta^d_o} = \bm{\Phi^d W_o^d}.
\end{aligned}
\end{equation}

The scatter matrix $\bm{S^d}$ encodes the local variance information at $d^{th}$ layer, and is given by:
\begin{equation}
\label{Eq:Sd_scatter}
\begin{aligned}
\bm{S^d} = \bm{\Phi^d\mathcal{L}^d(\Phi^d)^T}
\end{aligned}
\end{equation}

Now, by using (\ref{Eq:representation_theorem_oc}) and (\ref{Eq:Sd_scatter}), the minimization criterion in (\ref{Eq:ML-GOCELM}) is reformulated to the following: 
\begin{equation}\label{Eq:ML-GOCKELM1}
\begin{aligned}
\text{Minimize}:\pounds_{LMKOC^d}=\frac{1}{2}\bm{(W_o^d)^T(\Phi^d)^T(\Phi^d\mathcal{L}^d(\Phi^d)^T} \\ +\lambda \bm{I)\Phi^dW_o^d} + \frac{C}{2}\sum_{i=1}^{N}\left\|e_i^d  \right \|_2^{2}  \\
\text{Subject to}:\ \bm{(W_o^d)^T(\phi_i^d)^T\phi_i^d}=r - e_i^d, \text{  }i=1,2,...,N
\end{aligned}
\end{equation}

In addition, by substituting $\bm{K^d}=\bm{(\Phi^d)^T\Phi^d}$, where $\bm{k_i^d}\subseteq \bm{K^d}$, the optimization problem in (\ref{Eq:ML-GOCKELM1}) can be reformulated as follows:
\begin{equation}\label{Eq:ML-GOCKELM2}
\begin{aligned}
\text{Minimize}:\pounds_{LMKOC^d}=\frac{1}{2}\bm{(W_o^d)^T(K^d\mathcal{L}^d K^d}+\lambda \bm{K^d)W_o^d} \\ + \frac{C}{2}\sum_{i=1}^{N}\left\|e_i^d  \right \|_2^{2},  \\
\text{Subject to}:\ \bm{(W_o^d)^T k_i^d }= r - e_i^d, \text{  }i=1,2,...,N.
\end{aligned}
\end{equation}

The Lagrangian relaxation of (\ref{Eq:ML-GOCKELM2}) is shown below in (\ref{Eq:ML-GOCKELM2_lang}):	
\begin{equation}\label{Eq:ML-GOCKELM2_lang}
\begin{aligned}
\pounds_{LMKOC^d}=\frac{1}{2}\bm{(W_o^d)^T(K^d\mathcal{L}^d K^d}+\lambda \bm{K^d)W_o^d} \\ + \frac{C}{2}\sum_{i=1}^{N}\left\|e_i^d  \right \|_2^{2} - \sum_{i=1}^{N}\bm{\alpha_i^d}(\bm{(W_o^d)^T k_i^d } - r + e_i^d)
\end{aligned}
\end{equation}

where $\bm{\alpha^d = \{\alpha_i^d\}}, i=1,2 \hdots N$, is a Lagrangian multiplier. In order to optimize (\ref{Eq:ML-GOCKELM2_lang}), we compute its derivatives as follows:
\begin{equation}
\label{Eq:ML-GOCKELM2_deriv1}
\begin{aligned}
&\frac{\partial \pounds_{LMKOC^d}}{\partial \bm{W_o^d}} = 0 \Rightarrow \bm{W_o^d}=(\bm{\mathcal{L}^d K^d}+\lambda\bm{I})^{-1}\bm{\alpha^h}
\end{aligned}
\end{equation}
\begin{equation}
\label{Eq:ML-GOCKELM2_deriv2}
\begin{aligned}
&\frac{\partial \pounds_{LMKOC^d}}{\partial \bm{e_i^d}} = 0 \Rightarrow \bm{E^d}=\frac{1}{C}\bm{\alpha^h}
\end{aligned}
\end{equation}
\begin{equation}
\label{Eq:ML-GOCKELM2_deriv3}
\begin{aligned}
&\frac{\partial \pounds_{LMKOC^d}}{\partial \bm{\alpha_i^d}} = 0 \Rightarrow \bm{(W_o^d)^T K^d = r - E^h} \\
\end{aligned}
\end{equation}

The matrix $\bm{W_o^d}$ is obtained by substituting (\ref{Eq:ML-GOCKELM2_deriv2}) and (\ref{Eq:ML-GOCKELM2_deriv3}) into (\ref{Eq:ML-GOCKELM2_deriv1}), and is given by:
\begin{equation}
\label{Eq:W_ML-GOCKELM}
\begin{aligned}
\bm{W_o^d} &= \left(\bm{K^d}+\frac{1}{C}\bm{\mathcal{L}^dK^d}+\frac{\lambda}{C}\bm{I}\right)^{-1}\bm{r},
\end{aligned}
\end{equation}
$\bm{\beta^d_o}$ can be derived by substituting (\ref{Eq:W_ML-GOCKELM}) into (\ref{Eq:representation_theorem_oc}):
\begin{equation}
\label{Eq:ow_ML-GOCKELM}
\begin{aligned}
\bm{\beta_o^d} &= \bm{\Phi^d}\left(\bm{K^d}+\frac{1}{C}\bm{\mathcal{L}^dK^d}+\frac{\lambda}{C}\bm{I}\right)^{-1}\bm{r}.
\end{aligned}
\end{equation}

The predicted output of the final layer (i.e., $d^{th}$ layer) of the multi-layer architecture for training samples can be calculated as follows:
\begin{equation}
\label{Eq:oc_mlockelm}
\begin{aligned}
\bm{\widehat{O}= (\Phi^d)^T\beta_o^d=(\Phi^d)^T\Phi^d W_o^d=K^d(W_o^d)^T}
\end{aligned}
\end{equation}
where $\bm{\widehat{O}}$ is the predicted output for training data.

After completing the training process, a threshold is required to decide whether any sample is an outlier or not. Two types of threshold criteria ($\theta1$ and $\theta2$) are discussed in Subsection \ref{Sec:thresh}. 

The overall processing steps of $LMKOC$ is described in the Algorithm \ref{alg:ML-GOCKELM}.
 
\begin{algorithm}[!]
	\caption{$KRR$-based Multi-layer One-class classification with Local and Global Variance Information-based Embedding: $LMKOC$ and $GMKOC$}
	\label{alg:ML-GOCKELM}
	\begin{algorithmic}[1]
		\renewcommand{\algorithmicrequire}{\textbf{Input:}}
		\renewcommand{\algorithmicensure}{\textbf{Output:}}
		\REQUIRE  Training set $\bm{X}$, regularization parameter $(C)$, Graph regularization parameter $(\lambda)$, kernel function $(\Phi)$, number of layers $(d)$
		\ENSURE  Whether incoming sample is target or outlier \\
		\STATE Initially, $\bm{X^0=X}$
		\FOR {$h = 1$ to $d$}
		\IF{$h< d$}
		\STATE \textbf{First Phase}: First to $(d-1)^{th}$ layer stack Auto-Encoders in the hierarchical fashion for representation learning and transform the input $\bm{X^{h-1}}$.\\
		\IF{Local variance information-based embedding}
		\STATE Trained by $LKAE$ as per \ref{Eq:GKELM-AE2}
		\ELSIF{Global variance information-based embedding}
		\STATE Trained by $GKAE$ as per \ref{Eq:MVKELM-AE}
		\ENDIF
		\STATE Transformed output $\bm{X^{h}}$ for the input $\bm{X^{h-1}}$ is computed to pass as the input to the next layer in the hierarchy.
		\ELSE
		\STATE	\textbf{Second Phase}: Final layer i.e. $d^{th}$ layer for one-class classification. \\
		\STATE Output of $(d-1)^{th}$ Auto-Encoder is passed as an input to one-class classifier at $d^{th}$ layer. \\
		\IF{Local variance information-based embedding}
		\STATE Train the $d^{th}$ layer by $LMKOC^d$ as per \ref{Eq:ML-GOCKELM2}
		\ELSIF{Global variance information-based embedding}
		\STATE Train the $d^{th}$ layer by $GMKOC^d$ as per \ref{Eq:MV_ML-GOCELM}
		\ENDIF
		\ENDIF
		\ENDFOR
		\STATE Compute a threshold either $\theta1$ (\ref{Eq:Thr1}) or $\theta2$ (\ref{Eq:Thr2}).
		\STATE At final step, whether a new input is outlier or not, decides based on the rule discussed in \ref{Eq:df_ocelm2}.
	\end{algorithmic}
\end{algorithm}

\subsection{Global Variance Information based Graph-Embedded Multi-layer $KRR$ for One-class Classification: $GMKOC$ }\label{subsec:ML-MVOCKELM}
In this subsection, $GMKOC$ is proposed. In order to exploit global variance information for Auto-Encoder training, we define the variance $(\bm{Z^h})$ of the training data representations for the $h^{th}$ Auto-Encoder as follows:
\begin{equation}
\label{Eq:V_scatter1}
\begin{aligned}
\bm{Z^h} = \frac{1}{N}\sum_{i=1}^{N}(\bm{\phi_i^h}-\overline{\bm{\Phi^h}})(\bm{\phi_i^h}-\overline{\bm{\Phi^h}})^T,
\end{aligned}
\end{equation}
where $\overline{\bm{\Phi^h}}$ is the mean training vector in the kernel space of the $h^{th}$ Auto-Encoder, i.e. $\overline{\bm{\Phi^h}} = \frac{1}{N} \sum_{i=1}^N \bm{\phi_i^h}$. $\bm{Z^h}$ can be expressed in the form:
\begin{equation}
\label{Eq:V_scatter_kernel}
\begin{aligned}
\bm{Z^h} &= \frac{1}{N}\sum_{i=1}^{N}(\bm{\phi_i^h}-\overline{\bm{\Phi^h}})(\bm{\phi_i^h}-\overline{\bm{\Phi^h}})^T \\
&= \frac{1}{N}\bm{\Phi^h}(\bm{I}-\frac{1}{N}\bm{11}^T)(\bm{\Phi^h})^T \\
&= \bm{\Phi^h\mathcal{Z}^h(\Phi^h)}^T
\end{aligned}
\end{equation}
where, $\bm{1}\in\mathbb{R}^N$ is a vector of ones, $\bm{I}\in\mathbb{R}^{N\times N}$ is the identity matrix, and $\bm{\mathcal{Z}^h}$ represents Graph Laplacian matrix for $h^{th}$ layer. Any type of global variance based Laplacian Graph (e.g. Linear Discriminant Analysis ($LDA$) \cite{duda1973pattern} and Clustering-based Discriminant Analysis ($CDA$) etc.) can be exploited in the $GMKOC$.   


Minimization problems and their solutions for global variance case can be simply obtained from the equations of local variance (Section \ref{subsec:ML-GOCKELM}) by using $\bm{Z^h}$ instead of $\bm{S^h}$. Hence, optimization problem for Global variance information based $KAE$ ($GKAE$) is written as follows by using $\bm{Z^h}$ instead of $\bm{S^h}$ in (\ref{Eq:GELM-AE}):
\begin{equation}
\label{Eq:MVKELM-AE}
	\begin{aligned}
	\text{Minimize}:\pounds_{GKAE}=\frac{1}{2} Tr\Big(\bm{(\beta_a^h)^T(Z^h}+\lambda \bm{I)\beta_a^h} \Big) + \frac{C}{2}\sum_{i=1}^{N}\left\|\bm{e_i^h}  \right \|_2^{2}  \\
	\text{Subject to}:\ \bm{(\beta_a^h)^T \phi_i^h=x_i^{h-1} - e_i^h}, \text{  }i=1,2,...,N,
	\end{aligned}
\end{equation}

The use of (\ref{Eq:MVKELM-AE}) for the optimization of the proposed $GMKOC$, which minimizes the training error as well as class compactness simultaneously. This can be seen by expressing (\ref{Eq:MVKELM-AE}) using (\ref{Eq:V_scatter1}) as follows:
\begin{equation}
\label{Eq:Z_MVKELM-AE}
	\begin{aligned}
	\pounds_{GKAE} &=\frac{1}{2} Tr\Big(\bm{(\beta_a^h)^T(Z^h}+\lambda \bm{I)\beta_a^h} \Big)  + \frac{C}{2}\sum_{i=1}^{N}\left\|\bm{x_i^{h-1}-(\beta_a^h)^T \phi_i^h}  \right \|_2^{2}  \\
	&= \frac{1}{N}\sum_{i=1}^{N}\left ((\bm{(\beta_a^h)^T\phi_i^h}-\bm{(\beta_a^h)^T}\overline{\bm{\Phi^h}})^T(\bm{(\beta_a^h)^T\phi_i^h}-\bm{(\beta_a^h)^T}\overline{\bm{\Phi^h}})\right ) \\ & + \frac{C}{2}\sum_{i=1}^{N}\left\|\bm{x_i^{h-1}-(\beta_a^h)^T \phi_i^h}  \right \|_2^{2}  + \frac{\lambda}{2} Tr\Big(\bm{(\beta_a^h)}^T \bm{\beta_a^h}\Big)  \\
	&= \frac{1}{N}\sum_{i=1}^{N}\left\|\bm{o_i^h-o^h} \right \|_2^{2} + \frac{C}{2}\sum_{i=1}^{N}\left\|\bm{x_i^{h-1}-o_i^h} \right \|_2^{2} + \frac{\lambda}{2} Tr\Big(\bm{(\beta_a^h)}^T \bm{\beta_a^h}\Big)
	\end{aligned}
\end{equation}
where, $\bm{o_i^h} = \bm{(\beta_a^h)^T\phi_i^h}$ and $\bm{o^h} = \bm{(\beta_a^h)^T}\overline{\bm{\Phi^h}}$. Here, the regularization parameter $C$ provides the trade-off between the two objectives viz., minimizing the training error and class compactness.


Above minimization problem can be easily solved in a similar manner as solve the (\ref{Eq:GELM-AE}) in previous subsection. Hence, for global variance, we are providing only final solutions of the above minimization problems, due to space constraint, by using $\bm{\mathcal{Z}^h}$ instead of $\bm{\mathcal{L}^h}$ in (\ref{Eq:W_GKELM-AE}) and (\ref{Eq:ow_GKELM-AE}). The weights $\bm{W^h_a}$ and $\bm{\beta^h_a}$ for $GKAE$ are given by:
\begin{equation}
\label{Eq:W_MVKELM-AE}
\begin{aligned}
\bm{W^h_a} &= \left(\bm{K^h}+\frac{1}{C}\bm{\mathcal{Z}^h K^h}+\frac{\lambda}{C}\bm{I}\right)^{-1}\bm{X^{h-1}}
\end{aligned}
\end{equation}
\begin{equation}
\label{Eq:ow_MVKELM-AE}
\begin{aligned}
\bm{\beta^h_a} &= \bm{\Phi^h}\left(\bm{K^h}+\frac{1}{C}\bm{\mathcal{Z}^h K^h}+\frac{\lambda}{C}\bm{I}\right)^{-1}\bm{X^{h-1}}.
\end{aligned}
\end{equation}

After mapping the training data through the $(d-1)$ successive Auto-Encoder layers in the \textbf{first step}, the training data representations defined by the outputs of the $(d-1)^{th}$ $GKAE$ are used in order to train a \textbf{G}lobal variance based Graph-Embedded \textbf{M}ulti-layer \textbf{K}RR for \textbf{O}C\textbf{C} at $d^{th}$ layer ($GMKOC^d$) in the \textbf{second step}. Optimization problem of $GMKOC^d$ is written as follows by using $\bm{Z^d}$ instead of $\bm{S^d}$ in (\ref{Eq:ML-GOCELM}):
\begin{equation}
\label{Eq:MV_ML-GOCELM}
	\begin{aligned}
	\text{Minimize}:\pounds_{GMKOC^d}=\frac{1}{2}\bm{(\beta_o^d)^T(Z^d+\lambda I)\beta_o^d} + \frac{C}{2}\sum_{i=1}^{N}\left\|e_i^d  \right \|_2^{2}  \\
	\text{Subject to}:\ \bm{(\beta_o^d)^T \phi_i^d}=r - e_i^d, \text{  }i=1,2,...,N,
	\end{aligned}
\end{equation}


Above minimization problem can be solved similar as (\ref{Eq:ML-GOCELM}). Further, by using $\bm{\mathcal{Z}^d}$ instead of $\bm{\mathcal{L}^d}$ in (\ref{Eq:W_ML-GOCKELM}) and (\ref{Eq:ow_ML-GOCKELM}), its weight vectors $\bm{W_o^d}$ and $\bm{\beta^d_o}$ are obtained as follows:
\begin{equation}
\label{Eq:W_ML-MVOCKELM}
\begin{aligned}
\bm{W_o^d} &= \left(\bm{K^d}+\frac{1}{C}\bm{\mathcal{Z}^dK^d}+\frac{\lambda}{C}\bm{I}\right)^{-1}\bm{r},
\end{aligned}
\end{equation}
\begin{equation}
\label{Eq:ow_ML-MVOCKELM}
\begin{aligned}
\bm{\beta_o^d} &= \bm{\Phi^d}\left(\bm{K^d}+\frac{1}{C}\bm{\mathcal{Z}^dK^d}+\frac{\lambda}{C}\bm{I}\right)^{-1}\bm{r}.
\end{aligned}
\end{equation}

The predicted output of the final layer (i.e., $d^{th}$ layer) of the multi-layer architecture for training samples can be calculated as mentioned in (\ref{Eq:oc_mlockelm}) of previous subsection. The decision process for a test vector, whether it is outlier or not, is discussed in Subsection \ref{Sec:thresh}.

The overall processing steps followed by $GMKOC$ are described in Algorithm \ref{alg:ML-GOCKELM}. 

\subsection{Decision Function}\label{Sec:thresh}
Two types of thresholds namely, $\theta1$ and $\theta2$, are employed with the proposed methods, which are determined as follows:
\begin{enumerate}[1.]
	\item \textbf{For $\theta1$:}
	\begin{enumerate}[(i)]
		\item Calculate distance between the predicted value of the $i^{th}$ training sample and $r$, and store in a vector $\bm{d}$ as follows:
		\begin{equation}
		\label{Eq:dist1}
		\begin{aligned}	
		d(i)=\left |\widehat{O}_i - r\right |
		\end{aligned}
		\end{equation}
		\item After storing all distances in $\bm{d}$ as per (\ref{Eq:dist1}), sort these distances in decreasing order and denoted by a vector $\bm{d_{dec}}$. Further, reject few percent of training samples based on the deviation. Most deviated samples are rejected first because they are most probably far from the distribution of the target data. The threshold is decided based on these deviations as follows:
		\begin{equation}
		\label{Eq:Thr1}
		\begin{aligned}	
		\theta1=d_{dec}(\left \lfloor{\text{$\eta*N$}}\right \rfloor)\\
		\end{aligned}
		\end{equation}
		where $0<\eta\leq1$ is the fraction of rejection of training samples for deciding threshold value. $N$ is the number of training samples and $\lfloor\text{ } \rfloor$ denotes the floor operation.	
	\end{enumerate}
	\item \textbf{For $\theta2$:}
	Select threshold $(\theta2)$ as a small fraction of the mean of the predicted output:
	\begin{equation}
	\label{Eq:Thr2}
	\begin{aligned}	
	\theta2=(\left \lfloor{\text{$\eta*\text{mean}(\widehat{O})$}}\right \rfloor)\\
	\end{aligned}
	\end{equation}
	where $0<\eta\leq1$ is the fraction of rejection for deciding threshold value.
\end{enumerate}
So, a threshold value can be determined by above procedures. Afterwards, during testing, a test vector $\bm{{x}_p}$ is fed to the trained multi-layer architecture and its output $\widehat{O}_p$ is obtained. Further, compute $\widehat{d}$ for any one types  of threshold as follows: 

For $\theta1$, calculate the distance ($\widehat{d}$) between the predicted value $\widehat{O}_p$ of the $p^{th}$ testing sample and $r$ as follows:
\begin{equation}
\label{Eq:dist1_1}
\begin{aligned}	
\widehat{d}=\left |\widehat{O}_p - r\right |
\end{aligned}
\end{equation}

For $\theta2$, calculate the distance ($\widehat{d}$) between the predicted value $\widehat{O}_p$ of the $p^{th}$ testing sample and mean of the predicted values obtained after training as follows:
\begin{equation}
\label{Eq:dist2}
\begin{aligned}	
\widehat{d}=\left |\widehat{O}_p - \text{mean}(\widehat{O})\right |
\end{aligned}
\end{equation}

Finally, $\bm{{x}_p}$ is classified based on the following rule:
\begin{equation}
\label{Eq:df_ocelm2}
\begin{aligned}
&\text{If } \widehat{d} \leq \text{Threshold}, &\text{$\bm{{x}_p}$ belongs to normal class} \\
&\text{Otherwise}, &\text{$\bm{{x}_p}$ is an outlier}
\end{aligned}
\end{equation}

\section{Experimental Results}\label{Sec:performance}
In this section, experiments are conducted to evaluate the performance of the proposed MKOC over $21$ data sets. These datasets are obtained from University of California Irvine (UCI) repository \cite{Lichman:2013} and were originally generated for the binary or multi-class classification task. For our experiments, we have made it compatible with OCC task in the following ways. If a dataset has two or more than two classes then alternately, we use each of the classes in the dataset as the target class and the remaining classes as outlier class. In this way, we construct $21$ one-class datasets from $10$ multi-class datasets. Description of these datasets can be found in Table \ref{tab:Dataset}. These $21$ datasets can be divided into 3 category viz., $6$ financial, $8$ medical and $7$ miscellaneous datasets. Many of the datasets are slightly imbalanced. Class imbalance ratio of both of the classes are approximately $1:2$ in case of $11$ datasets viz., German($1$), German($2$), Pima($1$), Pima($2$), Glass(1), Glass($2$), Iono(1), Iono(2), Iris(1), Iris(2), and Iris(3). Here, all $7$ miscellaneous datasets are imbalanced in nature. All experiments on these datasets are carried out with MATLAB 2016a on Windows $7$ (Intel Xeon $3$ GHz processor, $64$ GB RAM) environment.

\begin{table}[tb]
	\centering
	\caption{Dataset Description}
	\begin{tabular}{|c|c|c|c|c|c|}
		\hline
		S. No. & Name  & \#Targets & \#Outliers & \#Features & \#samples \bigstrut\\
		\hline
		\multicolumn{6}{|c|}{\textbf{Financial Credit Approval Datasets}} \bigstrut\\
		\hline
		1     & Australia(1) & 307   & 383   & 14    & 690 \bigstrut[t]\\
		2     & Australia(2) & 383   & 307   & 14    & 690 \bigstrut[b]\\
		\hline
		3     & German(1) & 700   & 300   & 24    & 1000 \bigstrut[t]\\
		4     & German(2) & 300   & 700   & 24    & 1000 \bigstrut[b]\\
		\hline
		5     & Japan(1) & 294   & 357   & 15    & 651 \bigstrut[t]\\
		6     & Japan(2) & 357   & 294   & 15    & 651 \bigstrut[b]\\
		\hline
		\multicolumn{6}{|c|}{\textbf{Medical Disease Datasets}} \bigstrut\\
		\hline
		7     & Bupa(1) & 145   & 200   & 6     & 345 \bigstrut[t]\\
		8     & Bupa(2) & 200   & 145   & 6     & 345 \bigstrut[b]\\
		\hline
		9     & Ecoli(1) & 143   & 193   & 7     & 336 \bigstrut[t]\\
		10    & Ecoli(2) & 193   & 143   & 7     & 336 \bigstrut[b]\\
		\hline
		11    & Heart(1) & 160   & 137   & 13    & 297 \bigstrut[t]\\
		12    & Heart(2) & 137   & 160   & 13    & 297 \bigstrut[b]\\
		\hline
		13    & Pima(1) & 500   & 268   & 8     & 768 \bigstrut[t]\\
		14    & Pima(2) & 268   & 500   & 8     & 768 \bigstrut[b]\\
		\hline
		\multicolumn{6}{|c|}{\textbf{Miscellaneous Datasets}} \bigstrut\\
		\hline
		15    & Glass(1) & 76    & 138   & 9     & 214 \bigstrut[t]\\
		16    & Glass(2) & 138   & 76    & 9     & 214 \bigstrut[b]\\
		\hline
		17    & Iono(1) & 225   & 126   & 34    & 351 \bigstrut[t]\\
		18    & Iono(2) & 126   & 225   & 34    & 351 \bigstrut[b]\\
		\hline
		19    & Iris(1) & 50    & 100   & 4     & 150 \bigstrut[t]\\
		20    & Iris(2) & 50    & 100   & 4     & 150 \\
		21    & Iris(3) & 50    & 100   & 4     & 150 \bigstrut[b]\\
		\hline
	\end{tabular}%
	\label{tab:Dataset}%
\end{table}%

\subsection{Nomenclature of the Proposed and Existing Methods}\label{prop_exist}
Based on the multi-layer OCC described in the previous section, four variants have been proposed using two types of threshold criteria (viz., $\theta1$ and $\theta2$). Those variants are $LMKOC \mhyphen LLE\_\theta1$, $LMKOC \mhyphen LLE\_\theta2$, $GMKOC \mhyphen CDA\_\theta1$, and $GMKOC \mhyphen CDA\_\theta2$. Here, name of the used Laplacian graph and types of threshold criteria are concatenated with the name of the proposed methods. 

Total $11$ existing kernel-based one-class classifiers are employed for the comparison purpose, which can be categorized as follows:
\noindent
\begin{enumerate}[(i)]
	\item Support Vector Machine ($SVM$) based:  \textbf{O}ne-\textbf{c}lass \textbf{SVM} ($OCSVM$) \cite{scholkopf1999support}, \textbf{S}upport \textbf{V}ector \textbf{D}ata \textbf{D}escription ($SVDD$) \cite{tax1999support}
	\item $KRR$-based:
	\begin{enumerate}[(a)] 
		\item Without Graph-Embedding: $\textbf{K}RR$-based \textbf{O}C\textbf{C} ($KOC$) \cite{leng2014one} and $\textbf{K}RR$-based \textbf{A}uto-\textbf{E}ncoder model for \textbf{O}C\textbf{C} ($AEKOC$) \cite{gautam2017construction} 
		\item With Graph-Embedding: Two types of Graph-Embedding, i.e., Local and Global, have been explored in the literature. \textbf{L}ocal and \textbf{G}lobal Graph-Embedding with $KOC$ are named as $LKOC$-X \cite{iosifidis2016one} and $GKOC$-X \cite{iosifidis2016one,mygdalis2016one}, respectively. Here, X can be any Laplacian Graph with local or global Graph-embedding. For local, two types of Graphs are explored viz., \textbf{L}ocal \textbf{L}inear \textbf{E}mbedding ($LLE$) and \textbf{L}aplacian \textbf{E}igenmaps ($LE$). For global, four types of Graphs are explored viz., \textbf{L}inear Discriminant Analysis ($LDA$), \textbf{C}lustering-based L\textbf{DA} ($CDA$), class variance ($CV$), and sub-class variance ($SV$). Hence, final six existing variants are generated namely, $LKOC \mhyphen LE$\cite{iosifidis2016one}, $LKOC \mhyphen LLE$ \cite{iosifidis2016one}, $GKOC \mhyphen LDA$ \cite{iosifidis2016one}, $GKOC \mhyphen CDA$\cite{iosifidis2016one}, $GKOC \mhyphen CV$\cite{mygdalis2016one} and $GKOC \mhyphen SV$\cite{mygdalis2016one}. Here, we have considered the same Laplacian graphs as mentioned in \cite{iosifidis2016one}.
	\end{enumerate}
	\item Principal Component Analysis ($PCA$) based: \textbf{K}ernel \textbf{PCA} ($KPCA$)\cite{hoffmann2007kernel}.
\end{enumerate}
 
All existing and proposed one-class classifiers are implemented and tested in the same environment. $OCSVM$ is implemented using LIBSVM library \cite{CC01a}. $SVDD$ is implemented by using DD Toolbox \cite{Ddtools2015}. Codes of all $KRR$-based one-class classifiers were provided by the authors of the corresponding papers. The implementations of $KPCA$\cite{hoffmann2007kernel} and $AEKOC$\cite{gautam2017construction} are obtained from the links given in the paper (links are made available at the reference of the corresponding paper).

\subsection{Range of the Parameters of the Proposed and Existing Methods}\label{prop_exist_para}
For all of the kernel-based methods, Radial Basis Function (RBF) kernel is employed as shown below,
\begin{equation}\label{Eq:RBF_Kernel}
\begin{aligned}
\kappa(x_i, x_j) = exp\left ( -\frac{\left \| x_i-x_j \right \|_2^2}{2\sigma^2} \right )
\end{aligned}
\end{equation}
where $\sigma$ is calculated as the mean Euclidean distance between training vectors in the corresponding feature space. For the proposed multi-layer methods ($LMKOC$ and $GMKOC$), we have used maximum $d = 5$ layers and the value of $\sigma^h$ is calculated at each $h^{th}$ layer independently using the training data representations $X^{h-1}$. At each layer, regularization parameter is selected from the range of $\{2^{-3},\hdots,2^3\}$. The classifiers, which exploit graphs, have two regularization parameters, which are selected based on the cross-validation using values $2^l$, where $l=\{-3,...,3\}$. For the graph encoding subclass information in $GKOC \mhyphen SV$, the number of subclasses is selected from the range $\{2, 3,..., 20\}$. For $CDA$ graph-based classifiers ($GKOC \mhyphen CDA$, $GMKOC \mhyphen CDA\_\theta1$, and $GMKOC \mhyphen CDA\_\theta2$), number of clusters is selected from the range $\{2, 3,..., 20\}$. For the $KOC$ and $AEKOC$ methods, regularization parameter is selected from the range $\{2^{-3},\hdots,2^3\}$. For $KPCA$ based OCC, the percentage of the preserved variance is selected from the range $[85, 90, 95]$. The fraction of rejection $(\eta)$ of outliers during threshold selection is set equal to $0.05$ for all methods.

\subsection{Performance Evaluation Criteria}
Geometric mean ($\eta_g$) is computed in the experiment for evaluating the performance of each of the classifiers and is calculated as
\begin{equation}\label{Eq:Gmean}
\begin{aligned}
\eta_g = \sqrt{\text{Precision} * \text{Recall}}
\end{aligned}
\end{equation}

In all our experiments, $5$-fold cross-validation (CV) procedure is used  and the average Gmean value (along with the corresponding standard deviation ($\Delta$)) over $5$-fold CV are reported in the results. $\eta_g$ values of all of the classifiers are further analyzed by using mean of all Gmeans ($\eta_m$) and percentage of the maximum Gmean ($\eta_p$). $\eta_m$ is computed by taking average of all Gmeans obtained by a classifier over all datasets. $\eta_p$ is computed as follows \cite{fernandez2014we}:
\begin{equation}\label{Eq:PMG}
\eta_p=\frac{\sum_{i=1}^{\text{no. of datasets}}\left (\frac{\text{$\eta_g$ of a classifier for $i^{th}$ dataset}}{\text{Maximum $\eta_g$ achieved for $i^{th}$ dataset}} \times 100\right )}{\text{Number of datasets}}
\end{equation}
Moreover, Friedman testing is performed to verify the statistical significance of the obtained results. To this end, similar to \cite{fernandez2014we}, we also compute Friedman Rank ($\eta_f$)\cite{demvsar2006statistical} for ranking the classifiers.

\subsection{Performance Comparison}\label{perf_comp}
The Gmean ($\eta_g$) values of the $15$ kernel-based methods are provided in Table \ref{tab:perf_ab}-\ref{tab:perf_ef} for financial, medical, and miscellaneous datasets, respectively. Best $\eta_g$ per dataset is displayed in boldface in these Tables. 


\begin{table}[htbp]
	\tiny
	\centering
	\caption{Performance in terms of $\eta_g (\Delta)$ (\%) over 5-folds and 5 runs for financial datasets}
	\begin{adjustbox}{width=1.0\textwidth,center}
		\begin{tabular}{|l|c|c|c|c|c|c|}
		       \hline
		   One-class Classifiers & Australia(1) & Australia(2) & German(1) & German(2) & Japan(1) & Japan(2) \bigstrut\\
		   \hline
		   KPCA \cite{hoffmann2007kernel} & 63.69 (0.29) & 73.06 (0.18) & 80.77 (0.07) & 49.75 (0.28) & 64.09 (0.29) & 72.29 (0.29) \bigstrut\\
		   \hline
		   OCSVM \cite{scholkopf1999support} & 66.08 (0.6) & 76.59 (0.44) & 80.34 (0.33) & 52.8 (0.86) & 71.45 (0.38) & 75.78 (0.29) \bigstrut\\
		   \hline
		   SVDD \cite{tax1999support} & 65.55 (0.47) & 76.78 (0.22) & 81.1 (0.34) & 52.77 (0.82) & 70.15 (0.4) & 76.58 (0.28) \bigstrut\\
		   \hline
		   KOC \cite{leng2014one} & 65.07 (0.68) & 74.21 (1.12) & 73.17 (0.26) & 53.41 (0.3) & 67.33 (1.24) & 73.48 (1.62) \bigstrut\\
		   \hline
		   AEKOC \cite{gautam2017construction} & 72.88 (0.66) & 77.96 (0.55) & 74.04 (0.29) & 51.57 (0.37) & 76.23 (0.46) & 78.27 (0.51) \bigstrut\\
		   \hline
		   GKOC-LDA \cite{iosifidis2016one} & 64.98 (1.09) & 73.71 (1.21) & 72.53 (0.74) & 52.94 (0.24) & 67.12 (1.19) & 72.73 (1.83) \bigstrut\\
		   \hline
		   LKOC-LE \cite{iosifidis2016one} & 65.09 (0.82) & 74.03 (1.26) & 72.75 (0.5) & 53.06 (0.34) & 67.24 (1.2) & 73.15 (1.63) \bigstrut\\
		   \hline
		   LKOC-LLE \cite{iosifidis2016one} & 62.95 (0.64) & 70.72 (0.68) & 70.86 (0.52) & 52.51 (0.7) & 64.97 (0.39) & 69.79 (1.52) \bigstrut\\
		   \hline
		   GKOC-CDA \cite{iosifidis2016one} & 67.48 (1.67) & 73.74 (1.2) & 72.6 (0.66) & 52.9 (0.15) & 74.58 (2.56) & 72.75 (1.8) \bigstrut\\
		   \hline
		   GKOC-CV\cite{mygdalis2016one} & 63.21 (0.4) & 73.67 (0.51) & 81.42 (0.17) & 53.06 (0.08) & 63.91 (0.43) & 73.56 (0.36) \bigstrut\\
		   \hline
		   GKOC-SV \cite{mygdalis2016one} & 63.9 (0.67) & 74.6 (1.16) & 79.39 (1.63) & \textbf{53.49} (0.39) & 66.17 (0.69) & 73.96 (0.92) \bigstrut\\
		   \hline
		   LMKOC-LLE\_\theta1 & 77.81 (1.82) & 75.38 (1.3) & 74.04 (0.8) & 51.89 (0.61) & 80.8 (1.46) & 74.63 (1.55) \bigstrut\\
		   \hline
		   GMKOC-CDA\_\theta1 & 77.55 (1.74) & 79.55 (0.5) & 73.49 (0.71) & 53.37 (0.41) & 77.91 (2.56) & 79.52 (0.37) \bigstrut\\
		   \hline
		   LMKOC-LLE\_\theta2 & \textbf{81.7} (1.51) & 79.85 (0.67) & \textbf{82.19} (0.31) & 49.53 (0.86) & \textbf{83.72} (0.71) & 78.58 (0.49) \bigstrut\\
		   \hline
		   GMKOC-CDA\_\theta2 & 79.68 (0.97) & \textbf{80.49} (0.52) & 81.65 (0.54) & 49.56 (1.11) & 80.13 (0.59) & \textbf{80.72} (0.88) \bigstrut\\
		   \hline		   
		\end{tabular}%
	\end{adjustbox}
	\label{tab:perf_ab}%
\end{table}%


\begin{table}[htbp]
	\centering
	\caption{Performance in terms of $\eta_g (\Delta)$ (\%) over 5-folds and 5 runs for medical datasets}
		\begin{adjustbox}{width=1.0\textwidth,center}
	\begin{tabular}{|l|c|c|c|c|c|c|c|c|}
	    \hline
	One-class Classifiers & \multicolumn{1}{l|}{Bupa(1)} & \multicolumn{1}{l|}{Bupa(2)} & \multicolumn{1}{l|}{Ecoli(1)} & \multicolumn{1}{l|}{Ecoli(2)} & \multicolumn{1}{l|}{Heart(1)} & \multicolumn{1}{l|}{Heart(2)} & \multicolumn{1}{l|}{Pima(1)} & \multicolumn{1}{l|}{Pima(2)} \bigstrut\\
	\hline
	KPCA \cite{hoffmann2007kernel} & 62.91 (0.4) & 74.28 (0.59) & 65.79 (0.52) & 72.3 (0.44) & 70.42 (0.22) & 63.5 (0.78) & 77.98 (0.18) & 57.05 (0.4) \bigstrut\\
	\hline
	OCSVM \cite{scholkopf1999support} & 60.64 (1.3) & 69.78 (0.19) & 89.43 (0.93) & 79.42 (0.65) & 72.91 (0.55) & 64.9 (1.4) & 79.18 (0.19) & 56.59 (0.47) \bigstrut\\
	\hline
	SVDD \cite{tax1999support} & 60.64 (1.23) & 69.75 (0.21) & 89.49 (0.81) & 78.87 (0.45) & 72.91 (0.55) & 64.9 (1.4) & 79.21 (0.19) & 56.71 (0.62) \bigstrut\\
	\hline
	KOC \cite{leng2014one} & 57.09 (1.48) & 68.81 (0.99) & 89.38 (0.87) & 82.32 (0.39) & 65.03 (1.1) & 66.39 (0.53) & 79.04 (0.33) & 54.78 (0.21) \bigstrut\\
	\hline
	AEKOC \cite{gautam2017construction} & 56.19 (0.68) & 68.31 (0.53) & 89 (0.96) & 79.16 (1.38) & 67.99 (0.94) & 61.15 (0.68) & 78.66 (0.28) & 54.01 (0.39) \bigstrut\\
	\hline
	GKOC-LDA \cite{iosifidis2016one} & 57.04 (1.37) & 68.81 (1.04) & 88.79 (1.21) & 82.22 (0.85) & 64.44 (0.92) & 62.88 (0.8) & 78.94 (0.47) & 54.57 (0.3) \bigstrut\\
	\hline
	LKOC-LE \cite{iosifidis2016one} & 57.12 (1.66) & 68.77 (0.92) & 89.31 (1.15) & 82.28 (0.6) & 64.19 (0.96) & 64.38 (0.78) & 79.02 (0.33) & 54.58 (0.26) \bigstrut\\
	\hline
	LKOC-LLE \cite{iosifidis2016one} & 56.28 (0.79) & 67.72 (0.47) & 87.42 (0.61) & 82.24 (0.43) & 59 (1.14) & 66 (0.44) & 77.5 (0.68) & 52.02 (0.56) \bigstrut\\
	\hline
	GKOC-CDA \cite{iosifidis2016one} & 57.07 (1.38) & 68.81 (1.04) & 89.48 (0.77) & 82.8 (0.7) & 64.39 (0.74) & 64.84 (0.94) & 78.89 (0.48) & 54.5 (0.33) \bigstrut\\
	\hline
	GKOC-CV\cite{mygdalis2016one} & 62.78 (0.56) & 74.42 (0.9) & 86.03 (0.5) & 73.78 (0.46) & 69.02 (0.67) & 65.66 (0.4) & 77.56 (0.14) & 58.91 (0.4) \bigstrut\\
	\hline
	GKOC-SV \cite{mygdalis2016one} & 58.85 (1.45) & 70.07 (2.48) & 88.12 (4.53) & 86.08 (3.86) & 68.23 (2.18) & 67.28 (0.75) & 78.36 (0.37) & 55.59 (0.77) \bigstrut\\
	\hline
	LMKOC-LLE\_\theta1 & 61.69 (1.32) & 71.65 (0.51) & 89.13 (0.75) & 84.79 (1.82) & 67.53 (2.3) & 64.3 (2.83) & 78.2 (0.49) & 56.21 (0.83) \bigstrut\\
	\hline
	GMKOC-CDA\_\theta1 & 62.13 (1.79) & 72.83 (0.89) & \textbf{92.28} (0.34) & \textbf{86.32} (0.91) & 71.15 (0.68) & \textbf{69.86} (1.86) & 79.71 (0.69) & \textbf{58.92} (1.12) \bigstrut\\
	\hline
	LMKOC-LLE\_\theta2 & 62.04 (0.52) & 73.76 (0.41) & 88.66 (0.56) & 81.95 (1.33) & 70.28 (1.94) & 62.64 (3.95) & 80.15 (0.54) & 57.85 (0.76) \bigstrut\\
	\hline
	GMKOC-CDA\_\theta2 & \textbf{63.07} (0.86) & \textbf{75.31} (0.65) & 90.06 (1.06) & 82.77 (0.67) & \textbf{74.25} (0.94) & 58.79 (2.95) & \textbf{80.52} (0.38) & 58.44 (0.69) \bigstrut\\
	\hline
	\end{tabular}%
	\end{adjustbox}
	\label{tab:perf_cd}%
\end{table}%

\begin{table}[htbp]
	\centering
	\caption{Performance in terms of $\eta_g (\Delta)$ (\%) over 5-folds and 5 runs for miscellaneous datasets}
		\begin{adjustbox}{width=1.0\textwidth,center}
	\begin{tabular}{|l|c|c|c|c|c|c|c|}
    \hline
One-class Classifiers & Glass(1) & Glass(2) & Iono(1)  & Iono(2)  & Iris(1) & Iris(2) & Iris(3) \bigstrut\\
\hline
KPCA \cite{hoffmann2007kernel} & 57.69 (0.71) & 77.65 (0.21) & 76.54 (0.59) & 57.1 (0.23) & 96.44 (0.61) & 72.99 (1.37) & 69.29 (1.28) \bigstrut\\
\hline
OCSVM \cite{scholkopf1999support} & 59.61 (0.55) & 73.32 (0.88) & \textbf{93.13} (0.59) & 44.63 (0.51) & 85.06 (2.6) & 81.7 (2.29) & 83.18 (2.03) \bigstrut\\
\hline
SVDD \cite{tax1999support} & 59.61 (0.55) & 72.89 (0.69) & \textbf{93.13} (0.64) & 44.63 (0.51) & 84.12 (2.86) & 81.7 (2.29) & 82.67 (1.68) \bigstrut\\
\hline
KOC \cite{leng2014one} & 58.91 (1.18) & 73.08 (0.77) & 92.69 (0.23) & 53.4 (0.85) & 92.35 (0.87) & 85.59 (3.26) & 83.4 (2.04) \bigstrut\\
\hline
AEKOC \cite{gautam2017construction} & 59.29 (1.03) & 73.22 (0.77) & 92.95 (0.28) & 41.04 (0.52) & 92.79 (0.91) & 88.37 (1.89) & 83.74 (1.84) \bigstrut\\
\hline
GKOC-LDA \cite{iosifidis2016one} & 58.55 (1.73) & 73.14 (0.56) & 92.06 (0.43) & 43.04 (1.05) & 89.67 (0.63) & 84.34 (2.37) & 83.58 (2.84) \bigstrut\\
\hline
LKOC-LE \cite{iosifidis2016one} & 59.01 (1.37) & 72.94 (0.44) & 92.26 (0.38) & 45.64 (0.83) & 90.35 (0.07) & 85.03 (2) & 84.05 (2.42) \bigstrut\\
\hline
LKOC-LLE \cite{iosifidis2016one} & 59.23 (1.74) & 72.07 (1.3) & 89.18 (0.43) & 53.9 (1.64) & 92.35 (0.87) & 85.31 (2.48) & 80.83 (1.89) \bigstrut\\
\hline
GKOC-CDA \cite{iosifidis2016one} & 58.49 (1.75) & 73.14 (0.56) & 92.06 (0.28) & 43.04 (1.05) & 89.67 (0.63) & 85.68 (3.11) & 83.34 (2.61) \bigstrut\\
\hline
GKOC-CV\cite{mygdalis2016one} & 56.7 (1.41) & \textbf{79} (1.54) & 88.85 (0.65) & 57.15 (0.9) & 94.69 (1.01) & 80.76 (3.66) & 86.43 (3.92) \bigstrut\\
\hline
GKOC-SV \cite{mygdalis2016one} & 61.28 (3.36) & 74.43 (0.93) & 90.5 (1.53) & 55.32 (1.03) & 94.97 (2.53) & 87.42 (2.62) & 85.8 (2.43) \bigstrut\\
\hline
LMKOC-LLE\_\theta1 & 59.66 (2.29) & 76.59 (0.62) & 88.55 (0.56) & 60.3 (1.31) & \textbf{99.59} (0.56) & \textbf{90.26} (2.21) & \textbf{86.67} (2.43) \bigstrut\\
\hline
GMKOC-CDA\_\theta1 & \textbf{62.58} (1.74) & 76.84 (0.3) & 88.81 (0.86) & \textbf{67.5} (1.21) & 98.33 (1.41) & 76.95 (2.2) & 71.39 (1.64) \bigstrut\\
\hline
LMKOC-LLE\_\theta2 & 61.45 (2.07) & 78.45 (0.62) & 90.73 (1.12) & 49.28 (2.13) & 95.36 (1.64) & 90.9 (1.67) & 84.96 (5.55) \bigstrut\\
\hline
GMKOC-CDA\_\theta2 & 61.29 (2.07) & 78.23 (1.88) & 89.81 (0.8) & 65.02 (1.31) & 95.71 (2.38) & 78.79 (1.1) & 70.88 (2.07) \bigstrut\\
\hline
	\end{tabular}%
\end{adjustbox}
	\label{tab:perf_ef}%
\end{table}%

As per Table \ref{tab:perf_ab}, out of $6$ \textbf{financial} credit approval datasets, one of the proposed variants performs better than all $11$ existing methods in case of every dataset except German(2) dataset. For German(2) dataset, $LMKOC \mhyphen LLE\_\theta2$ exhibits comparable performance to $GKOC \mhyphen SV$. In case of Australian(1) dataset, all $4$ variants yield significantly (\textgreater $4\%$) better results compared to all of the methods presented in Table \ref{tab:perf_ab}. Explicitly,  $LMKOC \mhyphen LLE\_\theta2$ and $GMKOC \mhyphen CDA\_\theta2$ show improvement of $8.82\%$ and $6.8\%$, respectively, from the best $\eta_g$ value of the existing methods for Australian(1) dataset. For Australian(2), Japan(1) and Japan(2) datasets, best results obtained among all of the proposed methods exhibit significant difference of $2.53\%$, $7.49\%$, $2.45\%$, respectively, compared to the best $\eta_g$ obtained among all existing methods. Moreover, out of $6$ financial datasets, $LMKOC \mhyphen LLE\_\theta2$, $GMKOC \mhyphen CDA\_\theta2$ yield best $\eta_g$ for $3$ and $2$ datasets, respectively. 

As per Table \ref{tab:perf_cd}, out of $8$ \textbf{medical} datasets, one of the proposed variants performs better than all $11$ existing methods in case of every dataset. Moreover, $GMKOC \mhyphen CDA\_\theta1$ and $GMKOC \mhyphen CDA\_\theta2$, each yields best $\eta_g$ for $4$ datasets. For Ecoli(1) and Heart(2) datasets, $GMKOC \mhyphen CDA\_\theta1$ exhibits significant improvement of $2.79\%$ and $2.58\%$, respectively, from the best $\eta_g$ value of the existing methods.   

As we have discussed earlier, all $7$ \textbf{miscellaneous} datasets are imbalanced. Among $7$ miscellaneous datasets in Table \ref{tab:perf_ef}, one of the proposed variants performs better than all $11$ existing methods in case of every datasets except Glass(2) and Iono(1) datasets. Especially, for 3 datasets viz., Iono(2), Iris(1) and Iris(2) datasets, we obtain significant improvement of $10.36\%$, $3.15\%$, and $2.53\%$, respectively. In case of Glass(2) dataset, all $4$ proposed variants yield better result compared to all of the methods presented in Table \ref{tab:perf_ef} except $GKOC \mhyphen CV$ and $KPCA$.\\
\indent Overall, it can be observed from the above discussion and Table \ref{tab:perf_ab}-\ref{tab:perf_ef} that $GMKOC \mhyphen CDA\_\theta1$, and $GMKOC \mhyphen CDA\_\theta2$, $LMKOC \mhyphen LLE\_\theta2$, $LMKOC \mhyphen LLE\_\theta1$, $GKOC \mhyphen CV$, $GKOC \mhyphen SV$, $OCSVM$, and $SVDD$ yield best $\eta_g$ value for $6$, $6$, $4$, $2$, $1$, $1$, $1$ and $1$\footnote{ Here, $OCSVM$ and $SVDD$ yield best results for the same dataset i.e. Iono(1) dataset.} datasets, respectively. Hence, it can be stated that global variance-based embedding performs better compared to local variance-based embedding in most of the cases. Further, we compute $\eta_m$ and $\eta_p$ for all of the classifiers to analyze the $\eta_g$ value more closely.

\begin{figure}[t]
	\begin{center}
		\captionsetup{justification=centering}	
		\scalebox{0.65}[0.7]{\includegraphics{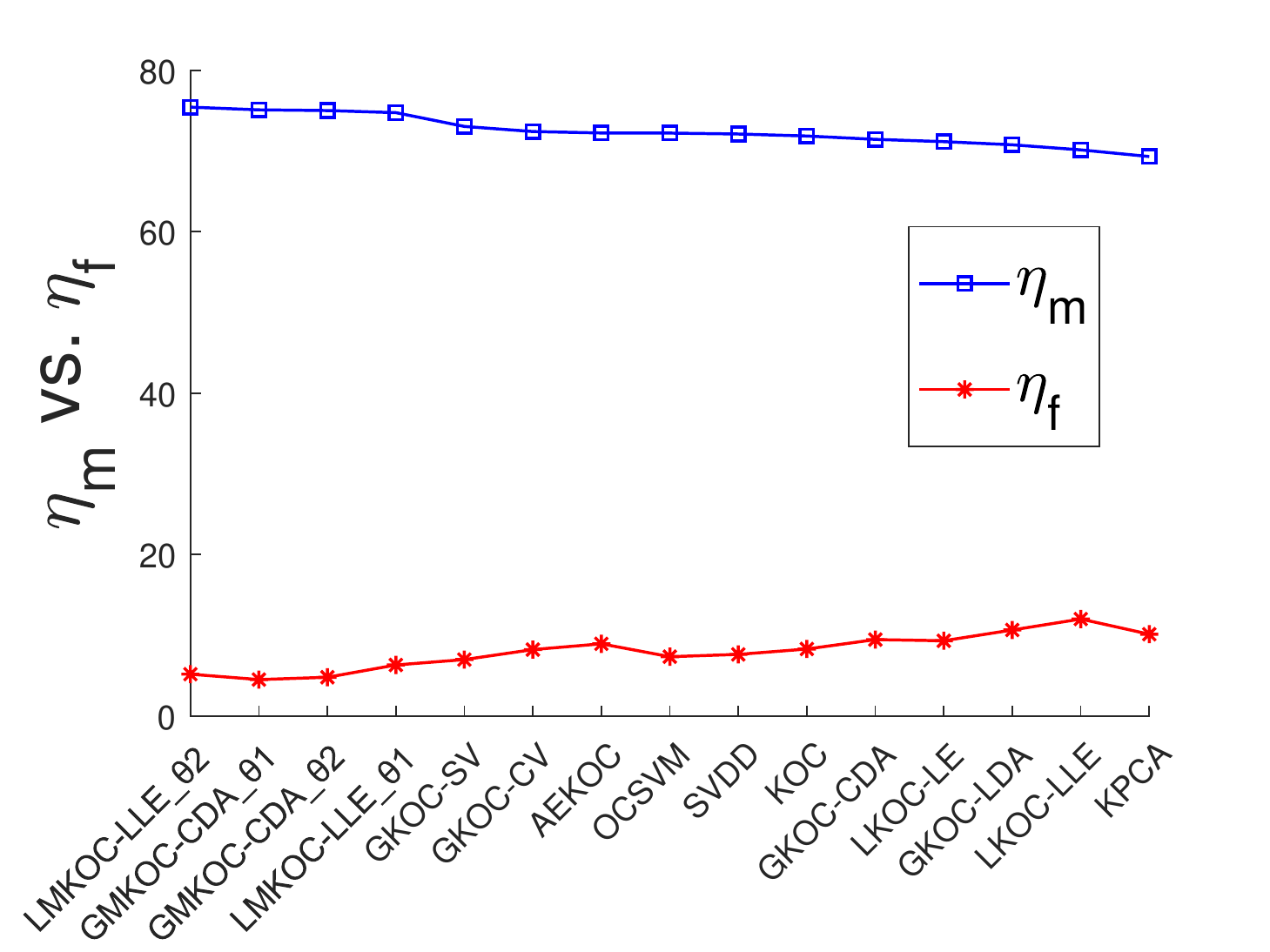}}
		\caption{All one-class classifiers as per $\eta_m$ in decreasing order and their corresponding Friedman Rank ($\eta_f$).}
		\label{fig:MGmean_FRank}
	\end{center}
\end{figure}%
\begin{table}[!h]
	\centering
	\caption{$\eta_f$ and $\eta_m$ of all one-class classifiers in increasing order of the $\eta_f$ (less value of $\eta_f$ indicates better performance).}
	\begin{tabular}{|l|c|c|}
		\hline
		\multicolumn{1}{|p{7.855em}|}{One-class Classifier} & \multicolumn{1}{p{4.215em}|}{$\eta_f$} & \multicolumn{1}{p{5.215em}|}{$\eta_m$(\%)} \bigstrut\\
		\hline
		GMKOC-CDA\_\theta1 & 4.52  & 75.10 \bigstrut[t]\\
		GMKOC-CDA\_\theta2 & 4.81  & 75.01 \\
		LMKOC-LLE\_\theta2 & 5.19  & 75.43 \\
		LMKOC-LLE\_\theta1 & 6.33  & 74.75 \\
		\hline \bigstrut[t]
		GKOC-SV & 7.00  & 73.04 \\
		OCSVM & 7.36  & 72.22 \\
		SVDD  & 7.64  & 72.10 \\
		GKOC-CV & 8.24  & 72.41 \\
		KOC   & 8.31  & 71.85 \\
		AEKOC & 8.95  & 72.23 \\
		LKOC-LE & 9.33  & 71.16 \\
		GKOC-CDA & 9.48  & 71.44 \\
		KPCA  & 10.14 & 69.31 \\
		GKOC-LDA & 10.67 & 70.77 \\
		LKOC-LLE & 12.02 & 70.14 \bigstrut[b]\\
		\hline		
	\end{tabular}%
	\label{tab:MGmean_FRank}%
\end{table}%

The performance of each method over $21$ datasets using $\eta_m$ metric is presented in Table \ref{tab:MGmean_FRank} and is plotted in a decreasing order in Fig. \ref{fig:MGmean_FRank}. $\eta_m$ metric provides average $\eta_g$ over $21$ datasets for a classifier. Based on the obtained results in Table \ref{tab:MGmean_FRank}, it can be clearly stated that all $4$ proposed variants, i.e., $LMKOC \mhyphen LLE\_\theta1$, $LMKOC \mhyphen LLE\_\theta2$, $GMKOC \mhyphen CDA\_\theta1$, and $GMKOC \mhyphen CDA\_\theta2$ have achieved top $4$ positions among $15$ one-class classifiers as per $\eta_m$ criterion. However, $GKOC \mhyphen SV$ yields best $\eta_m$ among existing kernel-based one-class classifiers. It is to be noted that $GMKOC \mhyphen CDA\_\theta1$ and $GMKOC \mhyphen CDA\_\theta2$ yield best $\eta_g$ for maximum number (i.e., $6$) of datasets, however, $LMKOC \mhyphen LLE\_\theta2$ emerges as the best classifier as per $\eta_m$ criterion. This is due to substantial improvement of $\eta_g$ for some of the datasets viz., Australia(1), Japan(1), and Iris(1). Hence, in order to further analyze the performance of the competing one-class classifiers, $\eta_p$ is calculated as per \ref{Eq:PMG}, similar to \cite{fernandez2014we}.

\begin{figure*}[!htbp]
	\begin{center}
		\resizebox{4.5in}{2.5in}{\includegraphics[width=0.75\textwidth]{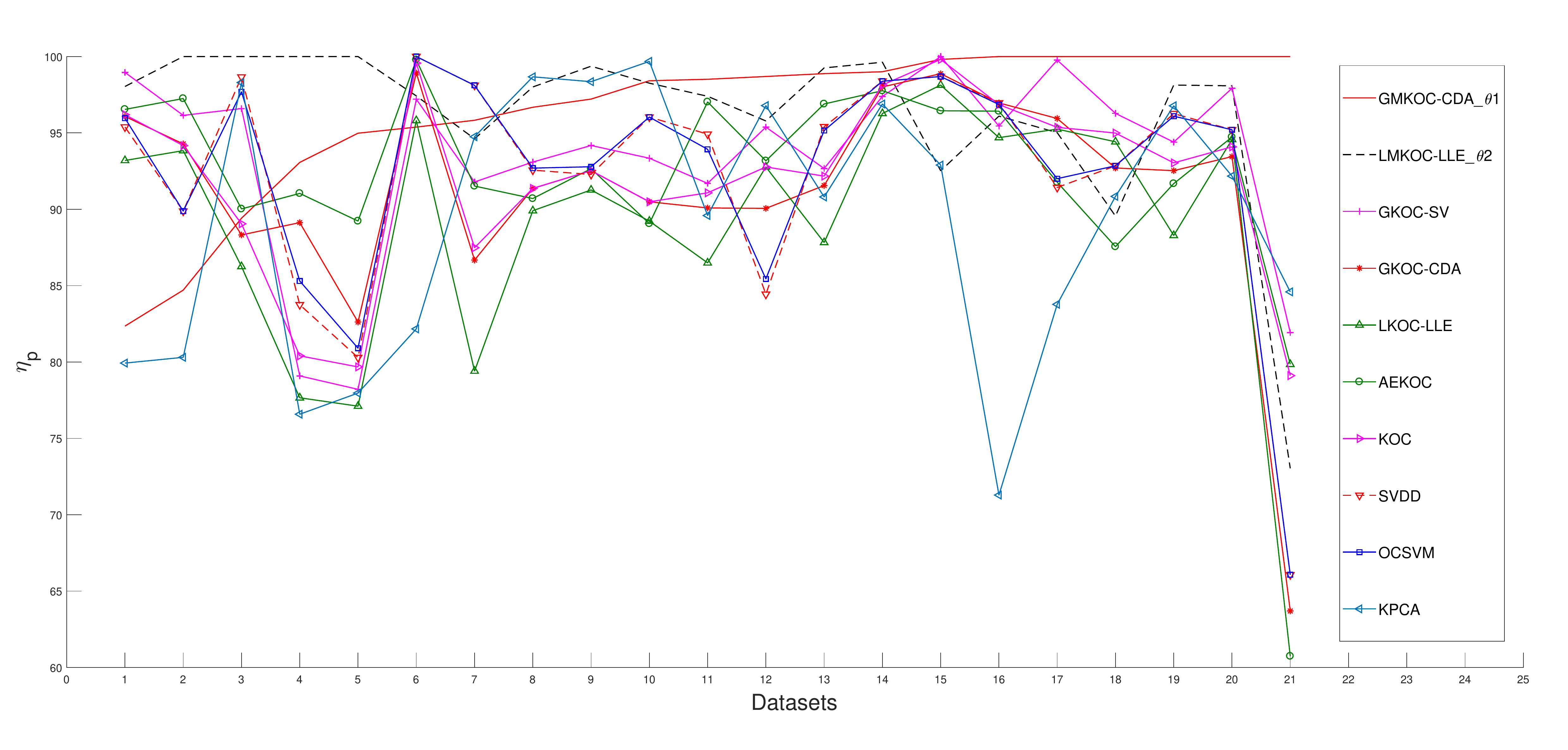}}
		\caption{$\eta_p$ achieved by various one-class classifiers over 21 datasets (ordered by increasing percentage)}
		\label{fig:PMG}
	\end{center}
\end{figure*}%

$\eta_p$ metric provides information regarding proximateness of each classifier towards maximum $\eta_g$ value. As it can be seen in Table \ref{tab:PMG}, $LMKOC \mhyphen LLE\_\theta1$, $LMKOC \mhyphen LLE\_\theta2$, $GMKOC \mhyphen CDA\_\theta1$, and $GMKOC \mhyphen CDA\_\theta2$ hold the top $4$ positions similar to the ranking based on the $\eta_m$ values in Fig. \ref{fig:MGmean_FRank}. It is to be noted that $GMKOC \mhyphen CDA\_\theta2$ yield best $\eta_g$ for $6$ datasets, and $LMKOC \mhyphen LLE\_\theta2$ for $4$ datasets, however, $LMKOC \mhyphen LLE\_\theta2$ yield better $\eta_p$ value compared to $GMKOC \mhyphen CDA\_\theta2$. It shows that indeed, $LMKOC \mhyphen LLE\_\theta2$ didn't yield best $\eta_g$ for maximum number of datasets but its $\eta_g$ values are more closer (compared to $GMKOC \mhyphen CDA\_\theta2$) to the best $\eta_g$ value of most of the datasets. In Fig. \ref{fig:PMG}, $\eta_p$ values of $10$ out of $15$ one-class classifiers are plotted in an increasing order for all of the datasets. All $15$ classifiers are not plotted for the sake clear visibility of the plotted lines. We have selected $10$ out of $15$ one-class classifiers based on the following discussion. Two out of four proposed variants, one global ($GMKOC \mhyphen CDA\_\theta1$) and one local variance-based ($LMKOC \mhyphen LLE\_\theta2$) multi-layer one-class classifiers, are selected to plot. Further, their corresponding single-layer one-class classifiers viz., $GKOC \mhyphen CDA$ and $LKOC \mhyphen LLE$, are also plotted. Out of two minimum class variance-based classifier ($GKOC \mhyphen SV$ and $GKOC \mhyphen CV$), $GKOC \mhyphen SV$ is plotted as it yields better $\eta_p$. Remaining all 5 one-class classifiers are also plotted with the above selected classifiers. 

\begin{table}[!b]
	\centering
	\caption{$\eta_p$ value over $21$ datasets}
	\begin{tabular}{|l|c|}
		\hline
		\multicolumn{1}{|p{7.855em}|}{One-class Classifiers} & \multicolumn{1}{p{8em}|}{$\eta_p$ (\%)} \bigstrut\\
		\hline
		GMKOC-CDA\_\theta1 & 96.33 \bigstrut[t]\\
		LMKOC-LLE\_\theta2 & 96.20 \\
		GMKOC-CDA\_\theta2 & 96.02 \\
		LMKOC-LLE\_\theta1 & 95.44 \\
		\hline \bigstrut[t]
		GKOC-SV & 93.41 \\
		GKOC-CV & 92.82 \\
		OCSVM & 92.38 \\
		SVDD  & 92.27 \\
		AEKOC & 92.00 \\
		KOC   & 91.83 \\
		GKOC-CDA & 91.16 \\
		LKOC-LE & 90.86 \\
		GKOC-LDA & 90.33 \\
		LKOC-LLE & 89.63 \\
		KPCA  & 89.20 \bigstrut[b]\\
		\hline
	\end{tabular}%
	\label{tab:PMG}%
\end{table}%

The plotted lines of the two single-layer ($GKOC \mhyphen CDA$ and $LKOC \mhyphen LLE$), and their corresponding multi-layer ($GMKOC \mhyphen CDA\_\theta1$ and $LMKOC \mhyphen LLE\_\theta2$) one-class classifiers in Fig. \ref{fig:PMG} clearly indicate the substantial performance improvement of the multi-layer version over single-layer one. Overall, Fig. \ref{fig:PMG} illustrates the clear superiority of the proposed multi-layer one-class classifiers over all $11$ existing methods. Moreover, $GMKOC \mhyphen CDA\_\theta1$ obtains more than $93\%$ $\eta_p$ value for all datasets except German(1), Iris(2), and Iris(3) datasets. Detailed $\eta_p$ values for all $15$ classifiers over $21$ datasets are made available on the link (\ULurl{https://goo.gl/QqUj4c}).

Above discussion suggests that all $4$ proposed variants emerge as the best performing classifier in terms of all employed performance evaluation criteria viz., $\eta_g$, $\eta_m$, and $\eta_p$. Despite this fact, a statistical testing needs to perform for verifying this fact. In the next subsection, Friedman Rank ($\eta_f$) testing is performed for statistical testing.

\subsection{Statistical Comparison}\label{sec:perf_FRank}
For comparing the performance of the $4$ proposed variants viz., $LMKOC \mhyphen LLE\_\theta1$, $LMKOC \mhyphen LLE\_\theta2$, $GMKOC \mhyphen CDA\_\theta1$, and $GMKOC \mhyphen CDA\_\theta2$, with the $11$ existing kernel-based methods on $21$ benchmark datasets, a non-parametric Friedman test is employed. In the Friedman test, the null hypothesis states that the mean of individual experimental treatment is not significantly different from the aggregate mean across all treatments and the alternate hypothesis states the other way around. Friedman test mainly computes three components viz., F-score, p-value and Friedman Rank ($\eta_f$). If the computed F-score is greater than the critical value at the tolerance level $\alpha=0.05$, then one rejects the equality of mean hypothesis (i.e. null hypothesis). We employ the modified Friedman test \cite{demvsar2006statistical} for the testing, which was proposed by Iman and Davenport \cite{iman1980approximations}. The F-score obtained after employing non-parametric Friedman test is $6.33$, which is greater than the critical value at the tolerance level $\alpha=0.05$ i.e. $6.33>1.72$. Hence, null hypothesis can be rejected with $95\%$ of a confidence level. The computed p-value of the Friedman test is $4.9414e-11$ with the tolerance value $\alpha=0.05$, which is much lower than $0.05$. This small value indicates that the differences in the performance of various methods are statistically significant. 

Afterwards, $\eta_f$ of each classifiers is also calculated to assign a rank to all $15$ one-class classifiers. Friedman test assigns a rank to all methods for each datasets. It assigns rank $1$ to the best performing algorithm, the second best rank $2$ and so on. If rank ties then average ranks are assigned \cite{demvsar2006statistical}. The $\eta_f$ values of all classifiers are provided in increasing order (less value of $\eta_f$ indicates better performance) in Table \ref{tab:MGmean_FRank}. These values are visualized in Fig. \ref{fig:MGmean_FRank} with the decreasing order of $\eta_m$. All $4$ proposed variants still achieve top four positions, similar to using the $\eta_m$ and $\eta_p$ metric. From Table \ref{tab:MGmean_FRank} and Fig. \ref{fig:MGmean_FRank}, it can be observed that $\eta_f$ of most of the classifiers follows a similar pattern as $\eta_m$, i.e., $\eta_f$ increases as $\eta_m$ decreases. However, some of the one-class classifiers don't follow the same pattern like $GKOC \mhyphen CV$ which has better $\eta_m$ but inferior $\eta_f$ compared to $OCSVM$ and $SVDD$. Among $4$ proposed variants, global variance-based methods ($GMKOC \mhyphen CDA\_\theta1$, and $GMKOC \mhyphen CDA\_\theta2$) outperform local-variance-based methods ($LMKOC \mhyphen LLE\_\theta1$, $LMKOC \mhyphen LLE\_\theta2$). Even, there is a significant difference ($1.52$) between the $\eta_f$ values of $GMKOC \mhyphen CDA\_\theta1$ and $LMKOC \mhyphen LLE\_\theta1$. The above analysis indicates that an one-class classifier with better $\eta_f$ value has better generalization scapability compared to the other existing methods.

Overall, after the performance analysis of all the $15$ one-class classifiers, it is observed that none of the existing one-class classifiers perform better than the proposed multi-layer one-class classifiers in terms of any discussed performance criteria.

\section{Conclusion}\label{Concl}
This paper has presented $4$ variants of Graph-Embedded multi-layer $KRR$-based one-class classifier. It is constructed by stacking various Graph-Embedded Auto-Encoders followed by a Graph-Embedded $KRR$-based one-class classifier. Stacked Graph-Embedded Auto-Encoder through multiple layers helps proposed classifiers in achieving better generalization and data representation capability. Overall, two types of training processes are involved i.e. one is for the Auto-Encoder and other is for the one-class classifier. We have explored two types of Graph-Embeddings, local and global variance-based embedding, in the kernel space of each layer using the Laplacian graph. $LLE$ and $CDA$ Laplacian graph are employed for local and global embedding, respectively. Extensive experimental comparisons have been provided with $11$ state-of-the-art kernel feature mapping based one-class classifiers over $21$ publicly available datasets in terms of $\eta_g$, $\eta_m$, $\eta_p$, and $\eta_f$. These experiments have exhibited that the proposed multi-layer one-class classifier provides state-of-the-art performance and outperformed all $11$ existing one-class classifiers. Moreover, the statistical significance of the results has also been verified by Friedman Ranking test. As per Friedman Rank, global variance-based proposed variants outperform local variance-based variants. In future work, various other types of available Auto-Encoder can be explored to enhance the performance of the proposed multi-layer architecture.

\vspace{1cm}
\noindent
\textbf{Funding Information}: This research was supported by Department of Electronics and Information Technology (DeITY, Govt. of India) under Visvesvaraya PhD scheme for electronics \& IT.

\section{Compliance with Ethical Standards}
\textbf{Conflict of Interest}: The authors declare that they have no conflict of interest.
\textbf{Ethical approval}: This article does not contain any studies with human participants or animals performed by any of the authors.

\bibliographystyle{unsrt}

\end{document}